\documentclass[a4paper,fleqn]{cas-dc}

\usepackage{subfig}
\usepackage{algorithm}
\usepackage{algpseudocode}
\usepackage[capitalize]{cleveref}
\usepackage[authoryear,longnamesfirst]{natbib}

\def\tsc#1{\csdef{#1}{\textsc{\lowercase{#1}}\xspace}}
\tsc{WGM}
\tsc{QE}
\tsc{EP}
\tsc{PMS}
\tsc{BEC}
\tsc{DE}

\begin{document}
\let\WriteBookmarks\relax
\def\floatpagepagefraction{1}
\def\textpagefraction{.001}

\shorttitle{PSE-Net}

\shortauthors{Shiguang Wang et~al.}

\title [mode = title]{PSE-Net: Channel Pruning for Convolutional Neural Networks with Parallel-subnets Estimator}
\tnotemark[1,2]

\tnotetext[1]{This work was supported by the National Natural Science Foundation of China (No.62071104, No. U2233209, No. U2133211, and No. 62001063).}

%
\author[1]{Shiguang Wang}[style=chinese]



\ead{xiaohu_wyyx@163.com}



\affiliation[1]{organization={ University of Electronic Science and Technology of China},
    addressline={No.2006, Xiyuan Ave, West Hi-Tech Zone}, 
    city={Chengdu},
    postcode={611731}, 
    country={China}}

\author[2]{Tao Xie}[style=chinese]

\author[3]{Haijun Liu}[style=chinese]
\ead{haijun_liu@126.com}
\author[4]{Xingcheng Zhang}[style=chinese]
\ead{zhangxingcheng@sensetime.com}

\affiliation[2]{organization={Harbin Institute of Technology},
    city={Harbin},
    postcode={150006}, 
    country={China}}
\affiliation[3]{organization={Chongqing University},
    city={Chongqing},
    postcode={400044}, 
    country={China}}
\affiliation[4]{organization={Sensetime Research},
    city={Beijing},
    postcode={100084}, 
    country={China}}    
\author%
[1]{Jian Cheng}[style=chinese]
\cormark[1]
\ead{chengjian@uestc.edu.cn}

\cortext[cor1]{Corresponding author}



\begin{abstract}
Channel Pruning is one of the most widespread techniques used to compress deep neural networks while maintaining their performances. 
Currently, a typical pruning algorithm leverages neural architecture search to directly find networks with a configurable width, the key step of which is to identify representative subnet for various pruning ratios by training a supernet.
However, current methods mainly follow a serial training strategy to optimize supernet, which is very time-consuming.
In this work, we introduce PSE-Net, a novel parallel-subnets estimator for efficient channel pruning.
Specifically, we propose a parallel-subnets training algorithm that
simulate the forward-backward pass of multiple subnets by
droping extraneous features on batch dimension, thus various subnets could be trained in one round. 
Our proposed algorithm facilitates the efficiency of supernet training and equips the network with the ability to interpolate the accuracy of unsampled subnets, enabling PSE-Net to effectively evaluate and rank the subnets. 
Over the trained supernet, we develop a prior-distributed-based sampling algorithm to boost the performance of classical evolutionary search. 
Such algorithm utilizes the prior information of supernet training phase to assist in the search of optimal subnets while tackling the challenge of discovering samples that satisfy resource constraints due to the long-tail distribution of network configuration. 
Extensive experiments demonstrate PSE-Net outperforms previous state-of-the-art channel pruning methods on the ImageNet dataset while retaining superior supernet training efficiency. 
For example, under 300M FLOPs constraint, our pruned MobileNetV2 achieves 75.2\% Top-1 accuracy on ImageNet dataset, exceeding the original MobileNetV2 by 2.6 units while only cost 30\%/16\% times than BCNet/AutoAlim.
\end{abstract}


\begin{highlights}
    \item{We propose PSE-Net, a novel parallel-subnets estimator for efficient channel pruning while evaluating the sampled subnets reliably. It's efficient and only needs 30\%/16\% total time compared with BCNet/AutoSlim for building the supernet estimator, respectively.}
    \item{We introduce the prior-distributed-based sampling algorithm, which facilitates the search of optimal subnets while resolving the challenge of discovering samples that fulfill resource constraints due to the long-tail distribution of network configuration.}
\end{highlights}

\maketitle
\section{Introduction}
Recently, Convolutional Neural Networks (CNNs) have evolved broader and achieved tremendous successes in various computer vision and machine learning applications, including image classification~\cite{krizhevsky2017imagenet, he2016deep}, object detection~\cite{redmon2016you, girshick2015fast}, semantic segmentation~\cite{lin2017refinenet, long2015fully}, and action recognition~\cite{wang2013action, herath2017going}. 
To enhance performance on these tasks, CNNs are becoming increasingly complex, with deeper and wider network designs. 
Nevertheless, deploying highly complex CNNs on modern devices with constrained resources, such as mobile devices, is impractical due to their excessive consumption of computational power and memory footprint. 
Many network compression approaches, including knowledge distillation~\cite{hinton2015distilling}, network pruning~\cite{guo2020dmcp,li2022revisiting,zhao2023exploiting,kang2019accelerator}, quantization~\cite{rastegari2016xnor}, and matrix decomposition~\cite{wang2017deepsearch} have been proposed to minimize the storage expenses and computation of a single CNN while retaining its performance. 
Among these methods, network pruning has been the most popular technique due to its simple implementation and high pruning ratio with minimal performance loss.

Given an unpruned network, network pruning seeks to produce a leaner network capable of precise assessment and fast inference time.   
Early pruning researches focus mostly on reducing individual weights with a negligible effect on the performance of deep neural networks, often known as weight pruning~\cite{zhang2018systematic, han2015learning}. 
Compared with weight pruning, channel pruning~\cite{guo2020dmcp,zhao2023exploiting,chen2022lap} is more straightforward since it eliminates the whole channel/filter rather than individual weights, making it possible to implement without the need for specialist software and hardware. 
Conventional channel pruning typically employs a pre-trained network and executes the pruning process end-to-end or layer-wisely. 
After pruning, it is deemed that a subnet with high accuracy is a plausible candidate for the ultimate solution. 

Recently, it has been advocated that the key to channel pruning should be learning a more compact network configuration, i.e., width, in lieu of the retained weights~\cite{liurethinking}. 
In this way, the quantity of filters or channels is utilized as a direct indicator of network width. 
Thus, subsequent researches employ neural architecture search (NAS) or other AutoML methods, including MetaPruning \cite{liu2019metapruning}, AutoSlim \cite{yu2019autoslim}, and TAS~\cite{dong2019network}, to directly identify the optimal network configuration. 
Typically, these methods entail three steps: \textbf{(1)} optimizing a \textbf{estimator (i.e., supernet)} over a number of epochs to effectively derive a benchmark performance estimator; 
\textbf{(2)} iteratively evaluating the trained supernet and obtaining the optimized \textbf{channel configurations (i.e., subnet)} under various resource constraints with minimal accuracy drop on the validation set; and \textbf{(3)} and retraining these optimized architectures from scratch under full training epochs. 

\begin{figure*}[]
    \centering
    \includegraphics[width=0.95\linewidth]{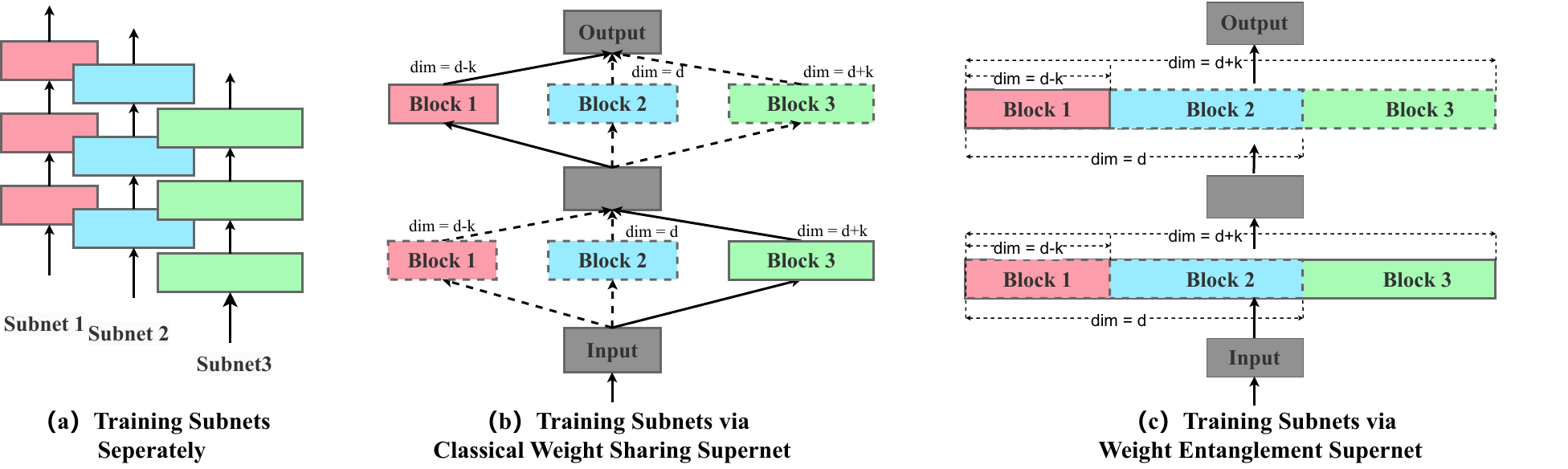}
    \caption{Different types of mechanisms for modeling estimator. (a). Thousands of distinct models are trained separately. (b) and (c) are two types of weight-sharing supernet mechanisms adopted by most one-shot NAS methods. Suppose we have three blocks with $d-k$, $d$, and $d+k$ channels, separately, the classical weight-sharing supernet (b) will separately build weights for these blocks, while the weight entanglement supernet will entangle these blocks with super weights.
    We denote the chosen parts in solid lines while the unchosen parts are in dashed lines. One-shot NAS methods like SPOS \cite{guo2020single}, FairNAS \cite{chu2021fairnas}, Cream \cite{peng2020cream} adopt the modeling mechanism (b), while AutoSlim \cite{yu2019autoslim}, BigNAS \cite{yu2020bignas}, BCNet \cite{su2021bcnet}, and Ours adopt the modeling mechanism (c).}
    \label{fig:weight_sharing}
    \vspace{-1em}
\end{figure*}
The most critical step of AutoML methods for channel pruning is training the benchmark performance estimator for evaluating various subnets. 
Theoretically, a decent estimator should adhere to two principle rules: \textbf{(1)} effectively identifying representative subnets under resource constraints and \textbf{(2)} efficiently evaluating different subnets. 
Customarily, one-shot NAS~\cite{guo2020single, peng2020cream} approaches have been used to train the estimator (i.e., supernet) due to their low computational overhead and competitive performance. 
As shown in \cref{fig:weight_sharing}, instead of training thousands of distinct models from scratch (\cref{fig:weight_sharing} \textcolor{red}{(a)}), one-shot approaches just train a single massive supernet (\cref{fig:weight_sharing} \textcolor{red}{(b)} and \cref{fig:weight_sharing} \textcolor{red}{(c)}), i.e., a performance estimator that can emulate any architectures in the search space, with weights shared among architectural candidates. 
Consequently, the performance of a network channel configuration refers to the precision of the defined subnet with shared supernet weights. 

However, sharing a single set of weights cannot ensure that each subnet receives sufficient training, hence diminishing the reliability of evaluating subnets. 
BCNet~\cite{su2021bcnet} characterizes such unreliability as unequal training periods for each channel in the convolution layer and proposes a new estimator, namely Bilaterally Coupled Network, to alleviate the training and assessment inequities. 
Nonetheless, the fairness of \emph{channel-level} training cannot fully reflect the corresponding \emph{network-level} training. 
In addition, BCNet is inefficient for training supernet as the training batch must be forwarded and backwarded twice at each training iteration. 
Alternatively, other approaches such as BigNAS~\cite{yu2020bignas} and AutoSlim \cite{yu2019autoslim} think about sufficient training from \emph{network-level}, which sample multiple subnets to execute forward-backward serially and aggregate the gradients from all sampled subnets before updating the weights of the supernet. It helps to provide more interpretive information by making the performance boundary of the search space available in validation, thereby improving the reliability of assessing subnets. 
However, such a process is roughly equivalent to training a plain model multiple times and thus slows the training efficiency of the supernet. 

In this work, we introduce PSE-Net, a novel parallel-subnets estimator for efficient channel pruning. 
Specifically, we follow the \emph{network-level} design paradigm that sampling multiple subnets for supernet updating at each iteration. To facilitate the supernet training, we propose the \textbf{parallel-subnets training algorithm}, a supernet training technique that focuses on time-efficient training by holding the supernet and dropping unnecessary features on batch dimension to simulate the forward-backward of multiple sampled subnets in one round. 
The parallel-subnets training algorithm enables the forward-backward execution of multiple networks in parallel, hence improving the training efficiency of the supernet while arming the network with the capability of interpolating the accuracy of unsampled subnets. 
In this way, after training, it is possible for PSE-Net to evaluate the performance of the sampled subnets reliably. 
As for the subsequent searching, we introduce the \textbf{prior-distributed-based sampling algorithm}, which draws architecture samples for performance accessing according to the prior distribution.
The \textbf{prior distribution} is efficiently estimated by utilizing the training loss and computation cost (e.g., FLOPs) of various subnets during supernet training. In which the training loss reflect the quality of each subnet while the computation cost represent the resource constraints.
Thus, to sample a random subnet subject to a FLOPs restriction, we directly employ rejection sampling from the prior distribution, which offers a significantly higher sampling efficiency and samping quality than directly sampling from the entire search space.

To summarize, the main contributions of this work are as follows:
\begin{itemize}
    \item{We propose PSE-Net, a novel parallel-subnets estimator for efficient channel pruning while evaluating the sampled subnets reliably. It's efficient and only needs 30\%/16\% total time compared with BCNet/AutoSlim for building the supernet estimator, respectively.}
    \item{We introduce the prior-distributed-based sampling algorithm, which facilitates the search of optimal subnets while resolving the challenge of discovering samples that fulfill resource constraints due to the long-tail distribution of network configuration.}
    \item{Comprehensive experiments on the ImageNet benchmark and downstream datasets indicate that our approach efficiently exceeds the SOTAs for a variety of FLOPs budgets. }
\end{itemize}

\section{Related works}
Network pruning techniques can be broadly categorized into two types, i.e., non-structured pruning and structured pruning. 
Non-structured pruning removes individual weights from the network, leading to a sparse weight matrix. 
However, this sparsity cannot be fully exploited in all the hardware implementations, thereby limiting the acceleration of the pruned network.
In comparison, structured pruning (i.e., channel pruning) removes whole filters, resulting in structure sparsity. This work is based on channel pruning and will sort out related works from the following two aspects. 

\subsection{Criterion Based Channel Pruning}
Numerous pruning techniques have been proposed to evaluate the importance of filters or channels in convolutional neural networks (CNNs).
Some approaches define specific criteria for assessing channel significance. 
For instance, \cite{lipruning, he2018soft} advocate for the use of the $l_1$-norm as a channel ordering criterion.
In a different vein, \cite{hu2016network} prunes channels based on a lower average proportion of zeros (APoZ) after ReLU activation.
Another strategy by \cite{molchanovpruning} involves approximating the change in the loss function through Taylor expansion, where channels causing the least change in the loss function are considered less significant.
Meanwhile, \cite{luo2017thinet} prunes filters based on statistical data from the subsequent layer, and \cite{he2019filter} computes the geometric average of filters within the same layer, excluding those closest to the average. \cite{lin2020hrank} suggests eliminating channels with low-rank feature maps, while \cite{he2020learning} employs diverse metrics to gauge the significance of filters across layers.
The GlobalPru, proposed by \cite{pan2023progressive} propose a channel attention-based learn-to rank framework to learn a global ranking of channels with respect to network redundancy.

These methods metrics-based prioritize the selection of "important" filters over discovering optimal channel topologies.
Despite the achievements,  the majority of these methods specify pruning proportions or hyperparameters in a hand-craft manner, which is complex, time-consuming, and makes it tough to identify optimal solutions.

\subsection{Search Based Channel Pruning}
Inspired by the advancements in neural architecture search (NAS) techniques~\cite{guo2020single, yu2019universally, yu2020bignas}, recent works resort to leveraging the NAS to identify optimal channel configurations. 
In \cite{he2018amc}, reinforcement learning is employed for layer-wise exploration of channel numbers.
Taking inspiration from curriculum learning, \cite{camci2022qlp} tailors pruning strategies for each sparsity level based on their complexity.
\cite{liu2019metapruning} and \cite{yu2019autoslim} firstly train a one-shot supernet and then seek efficient architectures inside it. 
\cite{dong2019network} introduces a differentiable NAS approach that can efficiently search transformable networks with variable width and depth. 
\cite{lin2021channel} provides a new approach for channel pruning based on the artificial bee colony algorithm (ABC), named ABCPruner, which seeks to identify optimal pruned structures in an efficient manner.
Characterizing channel pruning as a Markov process, \cite{guo2020dmcp} enables differentiable channel pruning.
Tackling biases in network width search, BCNet\cite{su2021bcnet} establishes a new estimator to alleviate training and assessment biases.

In either the process of training a large one-shot supernet or the process of searching for an ideal design, these approaches incur a substantial computational cost. 
To offer a efficient and realiable performance estimator, our poposed PSE-Net focuses on time-efficient training by holding the supernet and dropping unnecessary features on batch dimension to simulate multiple sampled subnets forward-backward in one round. 

\section{\large \bf Method}
\label{sec:methods}
We introduce our proposed method starting from the discussion of neural architecture search for channel pruning. Then, we will give the details of the proposed parallel-subnets training algorithm for efficient supernet training. 
Next, we introduce our prior-distributed-based sampling algorithm for efficient yet effective evolutional searching. 
Finally, we illustrate how to embed the proposed techniques into the pruning pipeline.
The overall framework is illustrated in \cref{fig:pipeline}.

\subsection{Neural Architecture Search for Channel Pruning}
\label{background}
Given an unpruned network $\mathcal{N}$, containing $L$ layers (e.g., convolutional layers and normalization layers), we can represent the architecture of $\mathcal{N}$ as $\mathcal{A}=\{\mathcal{C}_1, \mathcal{C}_2, ...,\mathcal{C}_l, ...,  \mathcal{C}_L\}$, where $\mathcal{C}_l$ is the output channel number of the $l$-th layer. 
Then, the objective of channel pruning is to layer-wisely make out redundant channels (indexed by $\mathcal{T}_l^{pruned}$), i.e., $\mathcal{T}_l^{pruned} \subset [1, \mathcal{C}_l]$, where $[1, \mathcal{C}_l]$ is an index set for all integers in the range of 1 to $\mathcal{C}_l$ for the $l$-th layer. 

However, recent empirical research by \cite{liurethinking} demonstrates that the absolute set of pruned channels $\mathcal{T}_l^{pruned}$ and their weights are not essential for the performance of a pruned network, but the resultant width $\mathcal{C}'_l$ of each layer is significant, i.e., 
\begin{equation}
    \mathcal{C}'_l = \mathcal{C}_l - |\mathcal{T}_l^{pruned}|.
\end{equation}

In this manner, recently popular AutoML-based structured pruning approaches view channel pruning as an automatic channel search with the {\bf search space $\mathcal{A}$}
amounts to $|\mathcal{A}| = \prod_{l=1}^{L}\mathcal{C}_l$, e.g., $20^{50}$ for $L=50$ layers and $\mathcal{C}_l=20$ channels. 
To reduce the search space, current methods tend to search at the group level rather than the channel level. 
Specifically, all channels at a layer are equally partitioned into $K$ groups, then we only need to consider $K$ cases: there are just $(\mathcal{C}_l/K)\cdot [1:K]$ layer width choices for the $l$-th layer. 
As a result, the search space is shrunk into $K^L$. 
Then, using neural architecture search (NAS) for channel pruning involves two phases: 

\textbf{Constraint-free Pre-training (phase 1):}
In this phase, we target learning the parameters of the weight-sharing supernet, i.e., the performance estimator. 
Specifically, the weights $\mathcal{W}$ of the target supernet $\mathcal{N}$ is trained by uniformly sampling a width $\mathcal{A}'$ and optimizing its corresponding subnet with weights $\mathcal{W}_{\mathcal{A}'} \subset \mathcal{W}$:
\begin{equation}
    \mathcal{W}^* = \min_{\mathcal{W}_{\mathcal{A}'} \in \mathcal{W}} \mathcal{L}_{trn}(\mathcal{A}', \mathcal{W}_{\mathcal{A}'}; \mathcal{N}, \mathcal{D}_{trn}),
    \label{eq:superne}
\end{equation}
where $\mathcal{L}_{trn}(\cdot)$ is the training loss function, $\mathcal{D}_{trn}$ represents the training dataset, and $\mathcal{W}^*$ denotes the trained weights.

\textbf{Resource-constrained Searching (phase 2):}
After the pre-training in phase 1, the optimal network width $\mathcal{A}^*$ corresponds to the one with the best performance on the validation dataset under the constraint resources, e.g., classification accuracy, 
\begin{equation}
\begin{aligned}
    \mathcal{A}^* = & arg \max_{\mathcal{A}' \in \mathcal{A}} Accuracy(\mathcal{A}', \mathcal{W}_{\mathcal{A}'}; \mathcal{W}, \mathcal{N}, \mathcal{D}_{val}), \\
    & s.t. \  Cons(\mathcal{N}_{pruned}(\mathcal{A}^{'}, \mathcal{W}^{'})) < \epsilon
    \label{min-val}
\end{aligned}
\end{equation}
where $\mathcal{D}_{val}$ denotes the validation dataset, $Cons(\cdot)$ is the proxy function for computational cost, $\epsilon$ is a specified budget of constraint resource. 
The searching process for \cref{min-val} can be performed efficiently by various algorithms, such as random search or evolutionary search. 
Then, the performance of the searched optimal width $\mathcal{A}^*$ is analyzed by training from scratch. 

\begin{figure*}[]
    \centering
    \includegraphics[width=0.95\linewidth]{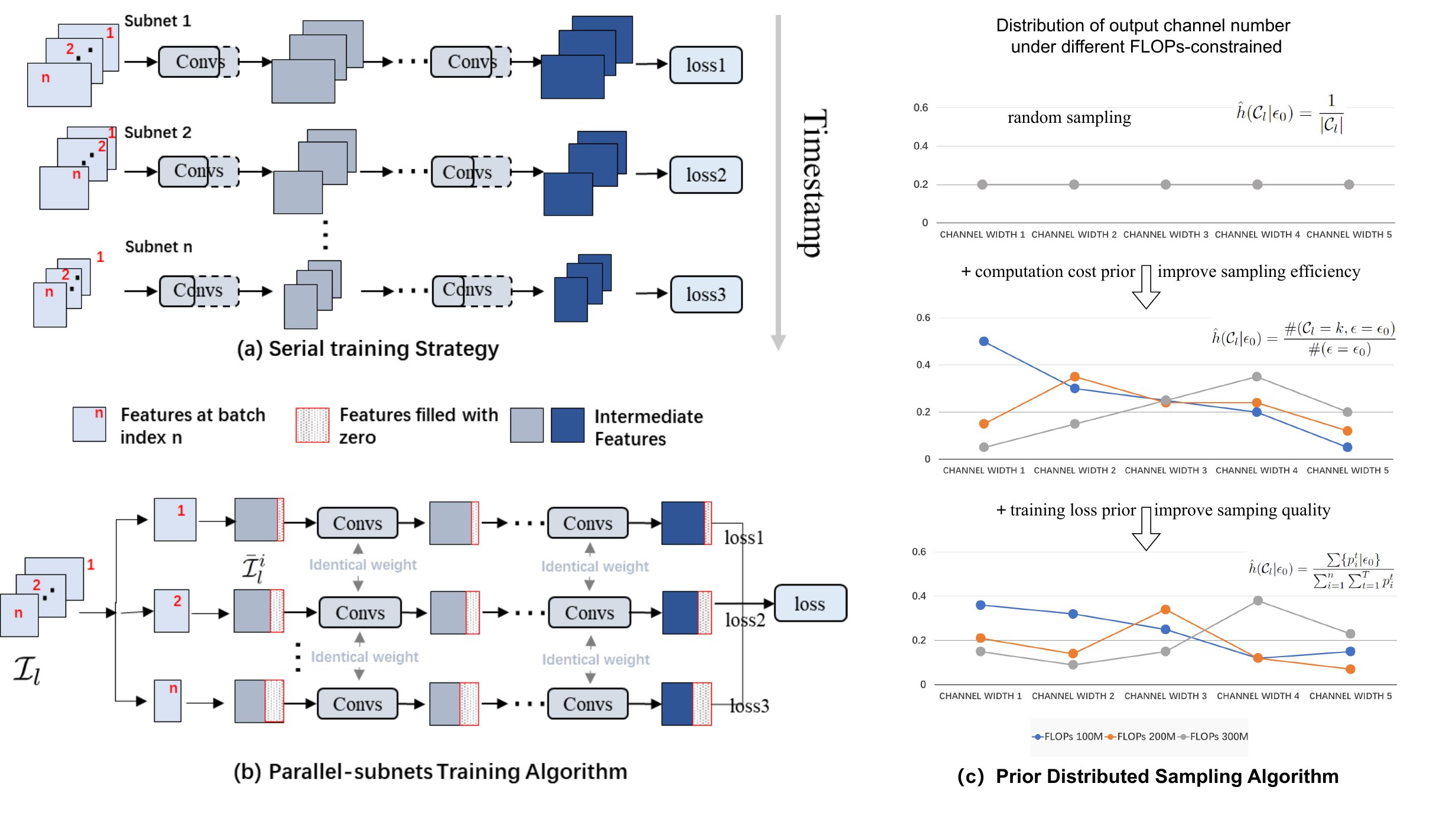}
    \caption{The overall framework of PSE-Net. We propose a parallel-subnets training algorithm and prior-distributed-based sampling algorithm to boost supernet training and subnet searching, respectively. Comparison of serial training strategy and parallel-subnets training algorithm during the supernet training phase is illustrated in (a) and (b). (a) Serial training strategy is approximately equal to training a plain model $n$ times, thus slowing the training efficiency of the supernet. (b) Parallel-subnets training algorithm that always holds the supernet and drops extraneous features on batch dimension to simulate multiple subnets forward-backward in one round. We denote the chosen parts in solid lines while the unchosen parts are in dashed lines. (c) Apart from introducing the computation cost for improving sampling efficiency, we also adopt the training loss that reflecting the quality of subnets to improve subnet sampling performance.}
    \label{fig:pipeline}
    \vspace{-1em}
\end{figure*}

The weight sharing strategy in \textbf{phase 1} reduces the search cost by several orders of magnitude. 
However, it still poses a potential problem, i.e., the insufficient training of subnets within the supernet. 
Due to such an issue, there is a limited correlation between the performance of an architecture measured by the one-shot weight and its actual performance. 
Therefore, the search based on the weight $\mathcal{W}$ may not uncover promising architectures, producing a less than ideal solution. 
Towards this end, some methods~\cite{yu2020bignas, yuslimmable} replace \cref{eq:superne} with following optimization problem: 
\begin{equation}
    \mathcal{W}^* = \min_{\mathcal{W}} \mathbb{E}_{\mathcal{A}' \in \mathcal{A}} [\mathcal{L}_{trn}( \mathcal{A}', \mathcal{W}_{\mathcal{A}'}; \mathcal{N}, \mathcal{D}_{trn})] + \gamma  \mathcal{R}(\mathcal{W}), 
    \label{eq:4}
\end{equation}
where $\mathcal{R}(\mathcal{W})$ is the regularization term, 
and $\mathbb{E}[\cdot]$ denotes the expectation of random variables. 

In practice, the expectation term in \cref{eq:4} is usually approximated by $n$ uniformly sampled architectures and solved by SGD optimization algorithm, formulated as: 
\begin{equation}
    \mathcal{W}^* = \min_{\mathcal{W}} \sum_{i=1}^{n}  \mathcal{L}_{trn}(\mathcal{A}_i',\mathcal{W}_{\mathcal{A}_i'}; \mathcal{N},  \mathcal{D}_{trn}). 
    \label{eq:5}
\end{equation}
Nevertheless, \textbf{such serial training strategy requires to execute forward-backward $n$ times and aggregate the gradients from all sampled models at each training iteration, which is approximately equal to training a plain model $n$ times, thus slowing the training efficiency of the supernet, as illustrated in \cref{fig:pipeline} \textcolor{red}{(a)}}. 
Thus, we propose parallel-subnets training algorithm that enables multiple networks to be executed forward-backward in parallel, improving the training efficiency of the supernet while arming the network with the capability of interpolating the accuracy of unsampled subnets. 

\subsection{From Serial Training Strategy to Parallel-subnets Training Algorithm}
\label{DPA}
For the $l$-th layer of $\mathcal{N}$, we define its weight as $w_l \in \mathbb{R}^{\mathcal{C}_{l} \times \mathcal{C}_{l-1} \times \mathcal{K}_l}$, where $\mathcal{K}_l$ is additional dimensions of the layer. 
The input feature maps of the $l$-th layer are denoted by $\mathcal{I}_l = \{\mathcal{I}_l^1, \mathcal{I}_l^2, ..., \mathcal{I}_l^p, ..., \mathcal{I}_l^{\mathcal{C}_{l-1}}\} \in \mathbb{R}^{b \times \mathcal{C}_{l-1} \times h_{l-1} \times w_{l-1}}$, where $b$ is the batch size of the input, and $h_{l-1}$ and $w_{l-1}$ are the height and width, respectively. 
For convenient portrayal, we denote the collection of weights connected to the $q$-th output channel as $w_l^{q,:}$, and the collection of weights connected to the $p$-th input channel as $w_l^{:,p}$. 
The $p$-th input feature map $I_l^p$ is connected to $w_l^{:,p}$ of the layer to generate the output feature map. 
The general operation of the $l$-th layer can be then defined as follows: 
\begin{equation}
    \mathcal{F}_l(\mathcal{I}_l) = \sum_{p=1}^{\mathcal{C}_{l-1}} \mathcal{I}_l^p \ast w_l^{:,p}
    \label{eq:6},
\end{equation}
where $\ast$ denotes the basic operator in CNN, such as 2D convolution and batch normalization. 

During the pruning process, given the input feature $\mathcal{I}_l = \{\mathcal{I}_l^1, \mathcal{I}_l^2, ..., \mathcal{I}_l^p, ..., \mathcal{I}_l^{\mathcal{C}'_{l-1}}\} \in \mathbb{R}^{b \times \mathcal{C}'_{l-1} \times h_{l-1} \times w_{l-1}}$, where $\mathcal{C}'_{l-1} \le \mathcal{C}_{l-1}$ is the pruned output channel of the $(l\raisebox{0mm}{-}1)$-th layer, each $p$-th feature map $\mathcal{I}_l^p$ has the same dimension with the others. 
In this way, for the $l$-th layer in the model $\mathcal{N}$, \cref{eq:6} can be expressed sequentially as the following formula: 

\begin{equation}
\left\{\begin{matrix}
 \mathcal{Y}^{q_1} = \mathcal{F}^{q_1}_l(\mathcal{I}_l) = \sum_{p=1}^{\mathcal{C}'_{l-1}} \mathcal{I}_l^p \ast w_l^{:q_1, p}\\
 \mathcal{Y}^{q_2} = \mathcal{F}^{q_2}_l(\mathcal{I}_l) = \sum_{p=1}^{\mathcal{C}'_{l-1}} \mathcal{I}_l^p \ast w_l^{:q_2, p}\\
 \mathcal{Y}^{q_i} = \mathcal{F}^{q_i}_l(\mathcal{I}_l) = \sum_{p=1}^{\mathcal{C}'_{l-1}} \mathcal{I}_l^p \ast w_l^{:q_i, p}\\ 
 \mathcal{Y}^{q_n} = \mathcal{F}^{q_n}_l(\mathcal{I}_l) = \sum_{p=1}^{\mathcal{C}'_{l-1}} \mathcal{I}_l^p \ast w_l^{:q_n, p},
\label{eq-seq}
\end{matrix}\right.
\end{equation}
where $q_i$, $i=1,2,...,n$ denotes the $i$-th random sampled output channel from the predefined search space, $\mathcal{Y}^{q_i} \in \mathbb{R}^{b \times \mathcal{C}_{l}^{q_i} \times h_{l} \times w_{l}}$ is the output feature by input features and the first $q_i$ channel weights, i.e., $w_l^{:q_i, p}$. 

Such naive training enables the simple training of numerous networks inside a single architecture without compromising performance. 
It helps to display more interpretive information by rendering the performance boundary of the search space accessible during validation. 
However, in each iteration, due to the different size of $w_l^{:q_i, p}$ in each subnet, $n$ subnets must run serially by taking the same input feature $\mathcal{I}_l$ as input, and then aggregate their gradients, which impacts the training efficiency of the supernet. 
Theoretically, it would slow down the training process $n$ times. 
Towards this end, \textbf{we propose a parallel-subnets training algorithm that leverages the largest network as the forward-backward proxy and simulates multiple sampled subnets along the batch dimension in one round, thereby accelerating the supernet training}, as shown in \cref{fig:pipeline} \textcolor{red}{(b)}. 
Specifically, we divide the input $\mathcal{I}_l$ into $n$ parts along the batch dimension, which is formulated as follows: 
\begin{equation}
    \mathcal{I}_l = [\mathcal{I}_l^1, \mathcal{I}_l^2, ..., \mathcal{I}_l^i, ..., \mathcal{I}_l^n], 
    \label{eq:7}
\end{equation}
where $\mathcal{I}_l^i \in \mathbb{R}^{\frac{b}{n} \times \mathcal{C}'_{l-1} \times h_{l-1} \times w_{l-1}}$ is the $i$-th part of $\mathcal{I}_l$, where each part corresponds to a subnet, respectively.

For the $i$-th part $\mathcal{I}_l^i =  \{\mathcal{I}_l^{1,i}, \mathcal{I}_l^{2,i}, ..., \mathcal{I}_l^{\mathcal{C}'_{l-1}, i}\} \in \mathbb{R}^{\frac{b}{n} \times \mathcal{C}'_{l-1} \times h_{l-1} \times w_{l-1}}$ of $\mathcal{I}_l$, we pad $\mathcal{I}_l^i$ to $\bar{\mathcal{I}}_l$ with 0 along the channel dimension so that it can be fed into the largest layer (the layer with the largest output channel): 
\begin{align}
    \bar{\mathcal{I}}_l^i = [\mathcal{I}_l^i, 0] = \{\mathcal{I}_l^{1,i}, \mathcal{I}_l^{2,i}, ..., \mathcal{I}_l^{\mathcal{C}'_{l-1}, i}, 0\},
    \label{eq:8}
\end{align}
where $\bar{\mathcal{I}}_l^i \in \mathbb{R}^{\frac{b}{n} \times C_{l-1} \times h_{l-1} \times w_{l-1}}$ is leveraged as the input of the largest layer. 
In this way, we can transform $\bar{\mathcal{I}}_l^i$ to the intermediate feature by using the largest layer, which is formulated as: 
\begin{equation}
    \bar{\mathcal{Y}}^i = \mathcal{F}_l(\bar{\mathcal{I}}_l^i) = \sum_{p=1}^{\mathcal{C}_{l-1}} \bar{\mathcal{I}}_l^{p,i} \ast w_l^{:,p},
    \label{eq:9}
\end{equation}
where $\bar{\mathcal{Y}}^i\in \mathbb{R}^{\frac{b}{n} \times \mathcal{C}_{l} \times h_{l} \times w_{l}}$ is the output of the largest layer $\mathcal{C}_l$. 

However, the $\bar{\mathcal{Y}}^i \in \mathbb{R}^{\frac{b}{n} \times \mathcal{C}_{l} \times h_{l} \times w_{l}}$ is not approximate to $\mathcal{Y}^{q_i} \in \mathbb{R}^{b \times \mathcal{C}_{l}^{q_i} \times h_{l} \times w_{l}}$ along the channel dimension. 
Supposing a function $\mathcal{G}(\cdot)$ that applies to $\bar{\mathcal{Y}}^{i}$, we can get $\mathcal{Y}^{q_i}$ through transformation: 
\begin{equation}
    \mathcal{Y}^{q_i} \propto \mathcal{G}(\bar{\mathcal{Y}}^i),
    \label{eq:10}
\end{equation}

Thus, by feeding $\bar{\mathcal{I}}_l^i, i=1,2,...,n$ into the largest convolutional layer at once, we get $\mathcal{Y}$ in one run as follows: 
\begin{equation}
    \mathcal{Y} = \mathcal{G}(\sum_{p=1}^{\mathcal{C}_{l-1}} [\bar{\mathcal{I}}_l^{p,1}, \bar{\mathcal{I}}_l^{p,2},...,\bar{\mathcal{I}}_l^{p,i},...,\bar{\mathcal{I}}_l^{p,n}]\ast w_l^{:,p}), 
    \label{eq:out}
\end{equation}
where $\mathcal{Y} = [\bar{\mathcal{Y}}^1, \bar{\mathcal{Y}}^2, ..., \bar{\mathcal{Y}}^n]$ equals $[\mathcal{Y}^{q_1}, \mathcal{Y}^{q_2}, ..., \mathcal{Y}^{q_n}]$ in the batch dimension if feeding the same data. 
In this manner, we are capable of imitating the concurrent training of many subnets in each mini-batch, enabling the network to interpolate the accuracy of unsampled subnets.
As such, it is possible to assess the performance of the sampled subnets in a reliable way after training. 
Next we derive the corresponding function $\mathcal{G}(\cdot)$ according to the common functions. 

\textbf{Convolution.}
We define a mask matrix $\mathcal{M}^i$ to enable $\bar{\mathcal{Y}}^i$ to approximate $\mathcal{Y}^{q_i}$ along the channel dimension so that the performance boundary of the search space is accessible during validation. 
Based on the linear property of convolution, we can get the ${(j,k)}$-th component of the mask matrix $\mathcal{M}^i$ as follows:
\begin{equation}
\mathcal{M}_{jk}^i = \begin{cases}
 1 \ \ \ if \ \frac{b}{n}\times i \le j < \frac{b}{n} \times (i+1), k \le \mathcal{C}_{l}^{q_i}, \\
 0\  \ \ otherwise
 \label{eq:M}
\end{cases}
\end{equation}
where $\mathcal{C}_{l}^{q_i}$ is the random sampled output channel of the $l$-th convolutional layer.  

Therefore, the function $\mathcal{G}(\cdot)$ for convolutional layer is element-wise product, which is formulated as:
\begin{equation}
    \mathcal{Y}^{q_i} \propto \mathcal{M}^i \otimes \bar{\mathcal{Y}}^i, 
    \label{eq:ele}
\end{equation}
where $\otimes$ is element-wise product. 
The linear layer can be viewed as a convolution layer with a convolution kernel of 1 and a stride of 1, hence its mask matrix is identical to that of the convolutional layer. 

{\bf Batch Normalization.} 
Typically, we often add a batch normalization layer after the convolution layer to expedite network training. 
Given the input feature $\mathcal{I}_l$, the pair of parameters $\gamma$ and $\beta$ scale and shift the normalized value as: $\gamma \mathcal{I}_l + \beta$.
These parameters are learned along with the original model parameters, and restore the representation capacity of the network. 
Based on this observation, we can see that batch normalization layer also has linear property. 
In this way, the function $\mathcal{G}(\cdot)$ has a same form with \cref{eq:M} and \cref{eq:ele}. 

Notably, batch norm statistics are not accumulated when training the supernet model, since they are ill-defined with varying architectures in parallel-subnets training algorithm. 
The bath norm statistics are re-calibrated after the training is completed, which is memory efficient and effective in handling weight-entanglement supernet training. 

\subsection{Searching Subnet with Prior-Distributed-Based Sampling Algorithm}
After the supernet $\mathcal{N}$ has been trained with $\mathcal{W}^*$, each channel configuration $\mathcal{A}'$ can be evaluated by estimating its performance (e.g., classification accuracy) on the validation dataset $\mathcal{D}_{val}$ using the \cref{min-val}. 
One of which critical steps entails drawing architecture examples according to diverse resource restrictions (e.g., FLOPs in this work). 
At each sampling step, it's necessary to first select a sample of target FLOPs $\epsilon$. 
Due to the inability of enumerating all evaluable structures, previous researchers resort to random search, evolutional search, and reinforcement learning to find the most promising architecture. 
However, \textbf{it may be hard to find samples that meet resource constraints, due to the long-tail distribution of network configuration}.
On the other hand, a network configuration that performs well on the training set also has a greater chance of performing better on the validation set.
If we collect prior information during the training of a supernet to construct a distribution, we can sample a network with superior performance more efficiently. 

Towards this end, we propose a novel search algorithm for promising architecture search in an efficient way. 
Specifically, we propose to approximate $h(\mathcal{A}|\epsilon_0)$ empirically to accelerate the FLOPs-constrained sampling procedure, where $h(\mathcal{A}|\epsilon_0)$ is the distribution over architectures conditioned on constraint $\epsilon_0$. 
As illustrated in \cref{DPA}, the supernet has a set of $L$ convolutional layers $\mathcal{A}=\{\mathcal{C}_1, \mathcal{C}_2, ..., 
\mathcal{C}_l, ...,\mathcal{C}_L\}$, where $\mathcal{C}_l$ denotes the output channel number of the $l$-th layer. 
Let $\hat{h}(\mathcal{A}|\epsilon_0)$ denote an empirical approximation of $h(\mathcal{A}|\epsilon_0)$, for simplicity, we relax, 
\begin{equation}
    \hat{h}(\mathcal{A}|\epsilon_0) \propto \prod_{l=1}^{L} \hat{h}(\mathcal{C}_l|\epsilon_0).
    \label{eq:Afen}
\end{equation}
Let $\#(\mathcal{C}_l=k, \epsilon=\epsilon_0)$ be the number of times that the pair $(\mathcal{C}_l=k, \epsilon=\epsilon_0)$ occurs in the sample pool of architecture-FLOPs. 
Thus, a straightforward method to approximate $\hat{h}(\mathcal{C}_l|\epsilon_0)$ is formulated as: 
\begin{equation}
    \hat{h}(\mathcal{C}_l|\epsilon_0) = \frac{\#(\mathcal{C}_l=k, \epsilon=\epsilon_0)}{\#(\epsilon=\epsilon_0)}.
    \label{eq:h2}
\end{equation}
However, \textbf{such a naive way does not consider the subnet quality during sampling, which can be reflected by the prior information during supernet training}. 

More specifically, as illustrated in \cref{fig:pipeline} \textcolor{red}{c}, during the supernet training, 
we sample various widths to generate subnets at each iteration, the quality of which can be measured by the training loss. 
Thus, we can capture numerous subnets with associated training loss and training iterations during supernet training. 
Notably, as our supernet forwards $n$ subnets in parallel at each iteration, we can acquire a loss collection of $n$ distinct subnets during each iteration, which is formulated as: 
\begin{equation}
    \mathcal{L}^t = \{
    \mathcal{L}_{trn}(\mathcal{W}_{\mathcal{A}^t_i}; \mathcal{A}^t_i)
    \}|_{i=1}^n,
    \label{eq:loss1}
\end{equation}
where $\mathcal{A}^t_i$ is the $i$-th sampled channel configuration at the $t$-th iteration during supernet training.

Suppose we train the supernet $T$ iterations. 
As a result, we can get the total loss collection at the end training of supernet:
\begin{equation}
    \mathcal{L} = \{\mathcal{L}^1, \mathcal{L}^2, ... , \mathcal{L}^t, ... , \mathcal{L}^T\}.
    \label{eq:loss2}
\end{equation}

Nevertheless, since the weight $\mathcal{W}$ of supernet is continuously updated, using the loss record over various iterations cannot accurately reflect the real potential of the corresponding channel configuration. 
Thus, we define proxy loss for each configuration as follows:
\begin{equation}
    p_i^t = \frac{\mathcal{L}_{trn}(\mathcal{W}_{\mathcal{A}_{max}^t};\mathcal{A}_{max}^t)}{\mathcal{L}_{trn}(\mathcal{W}_{\mathcal{A}_{max}^T};\mathcal{A}_{max}^T)} \mathcal{L}_{trn}(\mathcal{W}_{\mathcal{A}^t_i};\mathcal{A}^t_i),
    \label{eq:prox}
\end{equation}
where $\mathcal{A}_{max}^t$ is the largest subnet of the supernet, with $\mathcal{A}_{max}^t=\mathcal{A}_{max}^1=\mathcal{A}_{max}^2=...=\mathcal{A}_{max}^T$. 

Therefore, the empirical approximation of $\hat{h}(\mathcal{C}_l|\epsilon_0)$ can be estimated as: 
\begin{equation}
    \hat{h}(\mathcal{C}_l|\epsilon_0) = \frac{\sum\{p_i^t|\epsilon_0\}}{\sum_{i=1}^{n} \sum_{t=1}^{T} p_i^t}, 
    \label{eq:h1}
\end{equation}
where $\{p_i^t|\epsilon_0\}$ is the architecture collections that yield FLOPs $\epsilon_0$. 

\subsection{Pruning Pipeline}
\textbf{SuperNet Training.}
Since the whole unpruned network is composed of convolution layers, normalization layers, and activation layers, all of which possess linear properties, the unpruned network also retains linear properties, allowing our suggested parallel-subnets training algorithm to be applied layer-by-layer to the network. 
At each training iteration, we forward the supernet and drop unnecessary features along the batch dimension to simulate the forward-backward pass of multiple subnets in one round, thereby accelerating the training of the supernet and equipping it with the ability to interpolate the accuracy of unsampled subnets. 

It is worth noting that this does not require more GPU resources, as all one-shot NAS-based pruning algorithms require storing the entire supernet in memory. As for the computing memory overhead, other algorithms may also sample the largest subnet during the training process, and our algorithm always holds the supernet and drops extraneous features on batch dimension to simulate multiple subnets, so the maximum computational memory consumption is also the same.

\textbf{Prior-Distributed-Based Evolution Search.}
Over the trained supernet, we perform a prior-distributed-based evolution search to obtain the optimal subnets (channel configurations) under resource constraints. 
At the begining of the search, we pick $\frac{P}{2}$ architectures with the top $p_i^t$ and $\frac{P}{2}$ random architectures from the distribution $\hat{h}(\mathcal{A}|\epsilon_0)$ as seeds. 
The top $k$ architectures are chosen as parents for generating the next offspring by mutation and crossover.
At each generation of a crossover, two randomly selected candidates are crossed to produce a new individual. 
If the number of offspring produced by crossover and mutation is insufficient, we sample the architecture directly from $\hat{h}(\mathcal{A}|\epsilon_0)$. 

\textbf{Subnet Retrain.}
After obtaining the optimal subnets under resource constraints, we retrain them from scratch to derive their true performance.

\section{Experimental Results}
In this section, we conduct extensive experiments on the ImageNet dataset with various network architectures, e.g., ResNet50~\cite{he2016deep}, MobileNetV2~\cite{sandler2018mobilenetv2} and VGG~\cite{simonyan2014very}, to validate the efficacy of our PSE-Net. 

\subsection{Implementation Details}
\label{impl}
{\bf Search Space.}
For ResNet50/VGG, the original model is considered as the unpruned network for all models and the channels of each layer are partitioned into $K$=10/20 groups.
For MobileNetV2, we consider two types of unpruned networks: the original model (1$\times$ FLOPs) and a modified version with 1.5$\times$ FLOPs achieved through constant width scaling. The channels of each layer are partitioned into $K$=20 groups. 
We maintain a maximum pruning rate of 80\% per layer to ensure that the network performance does not degrade.

{\bf SuperNet Training.}
For ResNet50, the training recipe is set as: LAMB optimizer with weight decay 2e-2, initial learning rate 0.008, 120 training epochs in total. 
For MobileNetV2, the training recipe is set as: SGD optimizer with momentum 0.9, nesterov acceleration, weight decay 4e-5, and 90 training epochs in total. 
The initial learning rate for training supernet is first warming up from 0 to 0.8 within 4 epochs, then is reduced to 0 by cosine scheduler. 
For VGG, the training recipe is set as: SGD optimizer with momentum 0.9, weight decay 1e-4, the initial learning rate 0.1 decayed by 10x at the 30-$th$, 60-$th$ and 90-$th$ epoch, 100 training epochs in total. 
For all our experiments, we split the input into $n=4$ parts and simulate multiple networks (one largest subnet and three random middle subnets) to execute forward-backward pass in one round at each training iteration. 

{\bf Pruned Subnet Searching.}
The implementation of pruned subnet searching follows most protocols of evolution search in SPOS~\cite{guo2020single}. 
We set the number of seeds $P$ to 128, 
$k$ to 64, and mutation probability to 0.2. 

{\bf Pruned Subnet Retraining.}
We train all pruned models from scratch. 
The pruned models are trained on 16 Tesla A100 GPUs with a batch size of 2048. 
For a fair comparison, the pruned ResNet50/MobileNetV2/VGG is trained for 120/450/100 total epochs. 
The initial learning rate for training pruned MobileNetV2 is first warming up from 0 to 0.8 within 20 epochs, then is reduced to 0 by the cosine scheduler. 
Other settings are the same as those of the supernet training. 

\subsection{Results on ImageNet Dataset}
\begin{table}[]
\centering
\caption{Performance comparison of ResNet50~\cite{he2016deep} under ImageNet dataset. 
Performances are evaluated by using a single scale of 224$\times$224. 
FLOPs is computed under the input resolution of 224$\times$224.
\#Params denotes the number of parameters. 
 Methods marked with ``*” denote that reporting the results with knowledge distillation. $\dagger$ means the results reported in BCNet~\cite{su2021bcnet}.}
\label{resnet50}
\resizebox{0.48\textwidth}{!}{
\begin{tabular}{cccccc}
\toprule
\multicolumn{6}{c}{ResNet50}                                                                                                  \\ \midrule
\multicolumn{1}{c|}{FLOPs}          & \multicolumn{1}{c|}{Methods}     & FLOPs & \multicolumn{1}{c|}{\#Params} & Top-1  & Top-5  \\ \midrule
\multicolumn{1}{c|}{}                     & \multicolumn{1}{c|}{Uniform$^\dagger$}     & 3.0G  & \multicolumn{1}{c|}{19.1M}      & 75.9\% & 93.0\% \\
\multicolumn{1}{c|}{}                     & \multicolumn{1}{c|}{Random$^\dagger$}      & 3.0G  & \multicolumn{1}{c|}{-}          & 75.2\% & 92.5\% \\
\multicolumn{1}{c|}{}  & \multicolumn{1}{c|}{AutoSlim*~\cite{yu2019autoslim}}    & 3.0G  & \multicolumn{1}{c|}{23.1M}      & 76.0\% & -      \\
\multicolumn{1}{c|}{3G}                     & \multicolumn{1}{c|}{LEGR~\cite{chin2019legr}}        & 3.0G  & \multicolumn{1}{c|}{-}          & 76.2\% & -      \\
\multicolumn{1}{c|}{}                     & \multicolumn{1}{c|}{MetaPruning~\cite{liu2019metapruning}  \tiny{CVPR'19}} & 3.0G  & \multicolumn{1}{c|}{-}          & 76.2\% & -      \\
\multicolumn{1}{c|}{}                     & \multicolumn{1}{c|}{BCNet~\cite{su2021bcnet} \tiny{CVPR'21}}       & 3.0G  & \multicolumn{1}{c|}{22.6M}      & 77.3\% & 93.7\% \\ 
\multicolumn{1}{c|}{}                     & \multicolumn{1}{c|}{PEEL \cite{hou2024network} \tiny{PR'24}}   & 3.0G  & \multicolumn{1}{c|}{-}      & 76.6\% & -      \\
\multicolumn{1}{c|}{}                     & \multicolumn{1}{c|}{\textbf{PSE-Net}}       & \textbf{3.0G}  & \multicolumn{1}{c|}{\textbf{21.5M}}      & \textbf{77.9\%} & \textbf{93.7\%} \\ 
\midrule
\multicolumn{1}{c|}{}                     & \multicolumn{1}{c|}{Uniform$^\dagger$}     & 2.0G  & \multicolumn{1}{c|}{13.3M}      & 75.1\% & 92.7\% \\
\multicolumn{1}{c|}{}                     & \multicolumn{1}{c|}{Random$^\dagger$}      & 2.0G  & \multicolumn{1}{c|}{-}          & 74.6\% & 92.2\% \\
\multicolumn{1}{c|}{}                  & \multicolumn{1}{c|}{SSS~\cite{huang2018data} \tiny{ECCV'18}}         & 2.8G  & \multicolumn{1}{c|}{-}          & 74.2\% & 91.9\% \\
\multicolumn{1}{c|}{}                     & \multicolumn{1}{c|}{GBN~\cite{you2019gate} \tiny{NeurlPS'19}}         & 2.4G  & \multicolumn{1}{c|}{31.83M}     & 76.2\% & 92.8\% \\
\multicolumn{1}{c|}{}                     & \multicolumn{1}{c|}{TAS*~\cite{dong2019network} \tiny{NeurlPS'19}}        & 2.3G  & \multicolumn{1}{c|}{-}          & 76.2\% & 93.1\% \\
\multicolumn{1}{c|}{}                     & \multicolumn{1}{c|}{AutoSlim*~\cite{yu2019autoslim}}   & 2.0G  & \multicolumn{1}{c|}{20.6M}      & 75.6\% & -      \\
\multicolumn{1}{c|}{}                     & \multicolumn{1}{c|}{LEGR~\cite{chin2019legr}}        & 2.4G  & \multicolumn{1}{c|}{-}          & 75.7\% & 92.7\% \\
\multicolumn{1}{c|}{}                     & \multicolumn{1}{c|}{FPGM~\cite{he2019filter} \tiny{CVPR'19}}        & 2.4G  & \multicolumn{1}{c|}{-}          & 75.6\% & 92.6\% \\
\multicolumn{1}{c|}{2G}                     & \multicolumn{1}{c|}{MetaPruning~\cite{liu2019metapruning}  \tiny{CVPR'19}} & 2.0G  & \multicolumn{1}{c|}{-}          & 75.4\% & -      \\
\multicolumn{1}{c|}{}                     & \multicolumn{1}{c|}{DMCP~\cite{guo2020dmcp} \tiny{CVPR'20}}        & 2.2G  & \multicolumn{1}{c|}{-}          & 76.2\% & -      \\
\multicolumn{1}{c|}{}                     & \multicolumn{1}{c|}{BCNet~\cite{su2021bcnet} \tiny{CVPR'21}}       & 2.0G  & \multicolumn{1}{c|}{18.4M}      & 76.9\% & 93.3\% \\
\multicolumn{1}{c|}{}                     & \multicolumn{1}{c|}{Tukan et al. \cite{tukan2022pruning} \tiny{NeurlPS'22}}   & 2.0G  & \multicolumn{1}{c|}{-}      & 75.1\% & -      \\
\multicolumn{1}{c|}{}                     & \multicolumn{1}{c|}{SANP \cite{gao2023structural} \tiny{ICCV'23}}   & 1.9G  & \multicolumn{1}{c|}{-}      & 76.7\% & -      \\
\multicolumn{1}{c|}{}                     & \multicolumn{1}{c|}{Roy et al. 
\cite{miles2023reconstructing} \tiny{ICCVW'23}}   & 1.9G  & \multicolumn{1}{c|}{-}      & 75.6\% & -      \\
\multicolumn{1}{c|}{}                     & \multicolumn{1}{c|}{PEEL \cite{hou2024network} \tiny{PR'24}}   & 2.2G  & \multicolumn{1}{c|}{-}      & 76.5\% & -      \\
\multicolumn{1}{c|}{}                     & \multicolumn{1}{c|}{ARPruning \cite{yuan2024arpruning} \tiny{NN'24}}   & 1.8G  & \multicolumn{1}{c|}{11.0M}      & 72.3\% & -      \\
\multicolumn{1}{c|}{}                     & \multicolumn{1}{c|}{\textbf{PSE-Net}}       & \textbf{2.0G}  & \multicolumn{1}{c|}{\textbf{18.0M}}      & \textbf{77.3\%} & \textbf{93.4\%} \\\midrule
\multicolumn{1}{c|}{}                     & \multicolumn{1}{c|}{Uniform$^\dagger$}     & 1.0G  & \multicolumn{1}{c|}{6.9G}       & 73.1\% & 91.8\% \\
\multicolumn{1}{c|}{}                     & \multicolumn{1}{c|}{Random$^\dagger$}      & 1.0G  & \multicolumn{1}{c|}{-}          & 72.2\% & 91.4\% \\
\multicolumn{1}{c|}{}                     & \multicolumn{1}{c|}{AutoSlim*~\cite{yu2019autoslim}}   & 1.0G  & \multicolumn{1}{c|}{13.3M}      & 74.0\% & -      \\
\multicolumn{1}{c|}{}                     & \multicolumn{1}{c|}{MetaPruning~\cite{liu2019metapruning} \tiny{CVPR'19}} & 1.0G  & \multicolumn{1}{c|}{-}          & 73.4\% & -      \\
\multicolumn{1}{c|}{1G} & \multicolumn{1}{c|}{AutoPruner~\cite{luo2020autopruner} \tiny{PR'20}}  & 1.4G  & \multicolumn{1}{c|}{-}          & 73.1\% & 91.3\% \\
\multicolumn{1}{c|}{}                     & \multicolumn{1}{c|}{BCNet~\cite{su2021bcnet} \tiny{CVPR'21}}       & 1.0G  & \multicolumn{1}{c|}{12M}        & 75.2\% & 92.6\% \\
\multicolumn{1}{c|}{}                     & \multicolumn{1}{c|}{PEEL \cite{hou2024network} \tiny{PR'24}}   & 1.1G  & \multicolumn{1}{c|}{-}      & 74.7\% & -      \\
\multicolumn{1}{c|}{}                     & \multicolumn{1}{c|}{\textbf{PSE-Net}}       & \textbf{1.0G}  & \multicolumn{1}{c|}{\textbf{12.6M}}        & \textbf{75.2\%} & \textbf{92.0\%} \\
\bottomrule
\end{tabular}}
\end{table}

\begin{table}
\centering
\caption{Performance comparison of MobileNetV2 under ImageNet dataset. 
Performances are evaluated by using a single crop of 224$\times$224. 
FLOPs is computed under the input resolution of 224$\times$224.
\#Params denotes the number of parameters. 
 Methods marked with ``*” denote that reporting the results with knowledge distillation. $\dagger$ means the results reported in BCNet~\cite{su2021bcnet}.}
\label{mbv2}
\resizebox{0.48\textwidth}{!}{
\begin{tabular}{c|c|cc|cc}
\toprule
\multicolumn{6}{c}{MobileNetV2}\\
\midrule  
FLOPs & Methods & FLOPs & \#Params & Top-1 & Top-5\\
\midrule
\multirow{9}*{305M(1.5$\times$)}
&Uniform$^\dagger$&305M&3.6M&72.7\%&90.7\%\\
&Random$^\dagger$&305M&\text{-}&71.8\%&90.2\%\\
& AutoSlim*~\cite{yu2019autoslim} & 305M &5.7M &74.2\% &\text{-}\\
&BCNet~\cite{su2021bcnet} \tiny{CVPR'21}&305M&4.8M&73.9\%&91.5\%\\
&BCNet*~\cite{su2021bcnet} \tiny{CVPR'21}&305M&4.8M&74.7\%&92.2\%\\
&{PEEL \cite{hou2024network} \tiny{PR'24}}&300M&-&74.2\%&-\\
&{PEEL* \cite{hou2024network} \tiny{PR'24}}&300M&-&74.8\%&-\\
& \textbf{PSE-Net} & \textbf{305M} & \textbf{4.1M} & \textbf{74.2\%} & \textbf{91.7\%} \\
& \textbf{PSE-Net*} &\textbf{305M} & \textbf{4.1M} & \textbf{75.2\%} & \textbf{92.4\%} \\
\midrule
\multirow{13}*{200M}
&Uniform$^\dagger$&207M&2.7M&71.2\%&89.6\%\\
&Random$^\dagger$&207M&\text{-}&70.5\%&89.2\%\\
&AMC~\cite{he2018amc} \tiny{ECCV'18}&211M&2.3M&70.8\%&\text{-}\\
&AutoSlim*~\cite{yu2019autoslim}&207M&4.1M&73.0\%&\text{-}\\
&LEGR~\cite{chin2019legr}&210M&\text{-}&71.4\%&\text{-}\\
&MetaPruning~\cite{liu2019metapruning} \tiny{CVPR'19}&217M&\text{-}&71.2\%&\text{-}\\
&BCNet~\cite{su2021bcnet} \tiny{CVPR'21}&207M&3.1M&72.3\%&90.4\%\\
&BCNet*~\cite{su2021bcnet} \tiny{CVPR'21}&207M&3.1M&73.4\%&91.2\%\\
&SANP \cite{gao2023structural} \tiny{ICCV'23}&213M&3.6M&72.1\%&90.4\%\\
&{PEEL \cite{hou2024network} \tiny{PR'24}}&211M&-&73.0\%&-\\
&{PEEL* \cite{hou2024network} \tiny{PR'24}}&211M&-&73.9\%&-\\
&\textbf{PSE-Net} &\textbf{207M}&\textbf{3.0M}&\textbf{72.7\%}&\textbf{90.9\%}\\
&\textbf{PSE-Net*} &\textbf{207M}&\textbf{3.0M}&\textbf{73.5\%}&\textbf{91.4\%}\\
\midrule
\multirow{9}*{100M} 
&Uniform$^\dagger$&105M&1.5M&65.1\%&89.6\%\\
&Random$^\dagger$&105M&\text{-}&63.9\%&89.2\%\\
&MetaPruning~\cite{liu2019metapruning} \tiny{CVPR'19} &105M &\text{-} &65.0\% &\text{-}\\
&BCNet~\cite{su2021bcnet} \tiny{CVPR'21}&105M&2.3M&68.0\%&89.1\%\\
&BCNet*~\cite{su2021bcnet} \tiny{CVPR'21}&105M&2.3M&69.0\%&89.9\%\\
&{PEEL \cite{hou2024network} \tiny{PR'24}}&87M&-&66.0\%&-\\
&{PEEL* \cite{hou2024network} \tiny{PR'24}}&87M&-&66.6\%&-\\
&\textbf{PSE-Net} &\textbf{105M} &\textbf{1.8M} & \textbf{68.1\%} & \textbf{89.5\%} \\
&\textbf{PSE-Net*} &\textbf{105M} &\textbf{1.8M} &\textbf{69.4\%} & \textbf{90.3\%} \\
\bottomrule
\end{tabular}}
\end{table}

\begin{table}[!h]
\centering
    \caption{Performance comparison of VGG16 on ImageNet dataset. FLOPs is computed under the input resolution of 224$\times$224. \#Params denotes the number of parameters. }
    \resizebox{0.475\textwidth}{!}{
        \begin{tabular}{l|cccc}
            \toprule
              & FLOPs & \#Params  & Top-1     & Top-5      \\ \midrule
            VGG~\cite{simonyan2014very}       &  15.53G          &  138.37M                 & 73.74\% & 91.66\%       \\
            \textbf{PSE-Net}&\textbf{7.10G}&  \textbf{120.99M}& \textbf{71.72\%} & \textbf{90.62\%}  \\
            \textbf{PSE-Net}&\textbf{10.49G} &  \textbf{65.31M}&\textbf{71.45\%} & \textbf{90.43\%}       \\
              \bottomrule
    \end{tabular}}
    \label{VGG}
\end{table}

In this part, we compare our method with other multiple competing methods such as 
Roy et al. \cite{miles2023reconstructing},
PEEL \cite{hou2024network},
ARPruning \cite{yuan2024arpruning},
Tukan et al. \cite{tukan2022pruning},
SANP \cite{gao2023structural},
DMCP~\cite{guo2020dmcp}, TAS~\cite{dong2019network}, AutoSlim~\cite{yu2019autoslim}, 
MetaPruning~\cite{liu2019metapruning}, 
AMC~\cite{he2018amc}, LEGR~\cite{chin2019legr}, AutoPruner ~\cite{luo2020autopruner}, SSS~\cite{huang2018data}, and BCNet~\cite{su2021bcnet} on ImageNet dataset. 
Specifically, we search on the ResNet50 and MobileNetV2 with different FLOPs constraint, and report the accuracy on the validation dataset as~\cite{yu2019autoslim, guo2020dmcp}. 
Following BCNet~\cite{su2021bcnet}, we additionally take into account two naive baselines: (1) Uniform: we reduce the width of each layer by a constant factor to suit FLOPs budget requirements; (2) Random: we randomly pick 20 networks under the FLOPs restriction and train them for 50 epochs, and then continue training the network with the best performance and portray its final result alongside the results reported from BCNet~\cite{su2021bcnet}.  
In our experiments, the original ResNet50/MobileNetV2 possesses 25.5M/3.5M parameters and 4.1G/300M FLOPs with 77.6\%/72.6\% Top-1 accuracy, respectively. 

\textbf{Comparison with SOTA Methods on ResNet50.}
As shown in \cref{resnet50}, PSE-Net yields the best Top-1 accuracy on ImageNet by using ResNet50~\cite{he2016deep} as unpruned architecture. 
For example, comparing our 2G FLOPs ResNet50 to the original model, Top-1 accuracy reduces by only 0.3\%. 
Compared with the current state-of-the-art BCNet~\cite{su2021bcnet}, our 3G FLOPs and 2G FLOPs ResNet50 exceed it by 0.6\% and 0.4\% Top-1 accuracy. 
Besides, our 1G FLOPs ResNet50 achieves similar performance with BCNet under comparable FLOPs and parameters. 
Notably, PSE-Net is more efficient than BCNet in terms of supernet training, which is demonstrated in \cref{effe}.
Compared with AutoSlim~\cite{yu2019autoslim},
PSE-Net under all FLOPs levels surpasses it by non-trivial margins, demonstrating the efficacy of simulating multiple subnets forward-backward in one round. 
Moreover, compared with the two vanilla baselines and other previous works, PSE-Net also exhibits predominant performances. 

\textbf{Comparison with SOTA Methods on MobileNetV2.}
The pruned results for MobileNetV2 are reported in ~\cref{mbv2}. 
Specifically, our 300M FLOPs PSE-Net/PSE-Net* reaches 74.2\%/75.2\% Top-1 accuracy, which outperforms the state-of-the-art BCNet/BCNet*~\cite{su2021bcnet} by 0.3\%/0.5\% with comparable FLOPs and a decrease of 0.7M parameters. 
Compared with AutoSlim*~\cite{yu2019autoslim}, our 300M FLOPs PSE-Net* exceeds it by 1\%, further demonstrating the superiority of PSE-Net. 
In addition, our 200M FLOPs PSE-Net achieves 72.7\% Top-1 accuracy, which also exceeds BCNet~\cite{su2021bcnet}, MetaPruning~\cite{liu2019metapruning}, and LEGR~\cite{chin2019legr} by 0.4\%, 1.5\% and 1.3\% respectively. 
Compared with BCNet*~\cite{su2021bcnet} and AutoSlim*~\cite{yu2019autoslim}, our 200M FLOPs PSE-Net* also surpasses them by 0.1\% and 0.5 respectively. 
Compared to the two vanilla baselines, Uniform and Random, our all FLOPs level MobileNetV2 significantly outperforms both. 

\textbf{VGG Results on ImageNet Dataset.}
The pruned results for VGG are reported in ~\cref{VGG}. 
Our 7G FLOPs pruned VGG obtains 71.72\% top-1 accuracy while our 10G FLOPs pruned VGG derives 71.45\% top-1 accuracy, which is induced by the parameters of 7G FLOPs pruned VGG significantly higher than those of 10G FLOPs pruned VGG. 

\begin{figure*}[]
    \centering
    \includegraphics[width=0.97\textwidth]{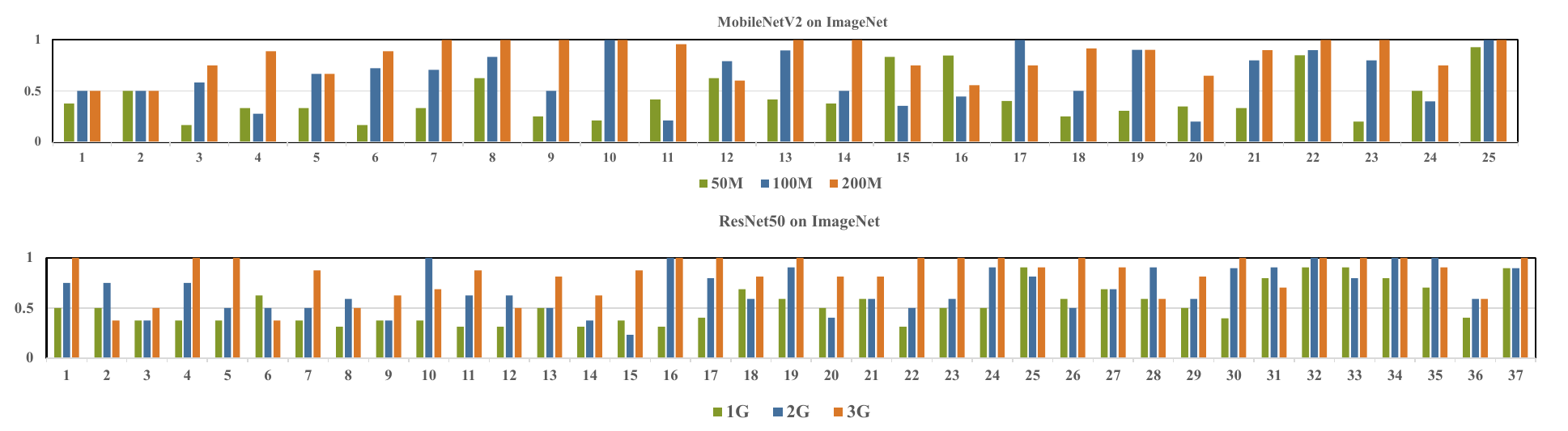}
    \caption{Visualization of the searched networks relative to different FLOPs. The abscissa represents the serial number of the convolutional layer. For each layer, the vertical axis represents the ratio of maintained channels to those of the original networks.}
    \label{fig:search}
    \vspace{-1em}
\end{figure*}

\textbf{Visualization.}
For intuitive comprehension, we visualize our searched PSE-Net under various FLOPs in \cref{fig:search}.  
In addition, the maintained ratio of layer widths relative to the original models is displayed for clarification. 
Yet, the last few layers have a greater likelihood of being retained. 
This might be due to the fact that the last few layers are more sensitive to the classification performance, and hence, it can be securely maintained when the FLOPs are decreased, as the same phenomenon is observed in BCNet~\cite{su2021bcnet}. 
          
\subsection{Effect and Efficiency Analysis of Parallel-subnets Training Algorithm}
\label{effe}
In this part, we compare the effect and efficiency of the parallel-subnets training algorithm with prior works by leveraging ResNet50 as unpruned architecture.  We construct ablation experiments as follows:
(1) SPOS$^\ddag$: we train the supernet by randomly sampling a subnet at each training iteration, which is also adopted by SPOS~\cite{guo2020single}.
(2) BCNet: at each training iteration, following~\cite{su2021bcnet}, two bilaterally coupled subnets are sampled for supernet training, where the selected channel width sums up to the max channels in the same layer.
(3) AutoSlim$^\ddag$: we sample 4 subnets, i.e., the max subnet, the min subnet, and two random sampled subnets. Data is serially fed into the subnets and their gradients are accumulated for parameters updating.
The sandwich rule is widely adopted in~\cite{yu2019autoslim,yu2020bignas,wang2021attentivenas}. 
For effect analysis, we train the supernet by 120 epochs by using 16 Tesla A100 GPUs following the training recipe of PSE-Net. For efficiency analysis, an AutoML-based pruning method typically involves three steps: Step1: Optimizing a supernet over a number of epochs to effectively derive a benchmark performance estimator; Step2: Iteratively evaluating the trained supernet and selectively reducing the channels with minimal accuracy drop on the validation set; and Step3: Obtaining the optimized channel configurations under various resource constraints and retraining these optimized pruning configurations. For Step3, all the methods share almost the same pipeline. For Step2, the main cost is incurred in evaluating accuracy on the validation set, we kept the number of channel configurations evaluated the same for fair comparison. In Step1, since different methods use different GPUs to training their supernet, we compare the number of executing forward-backward times, time cost at each training iteration and total training epochs. The iter\_time is measured on a Tesla A100 GPU with input shape of data as (128, 3, 224, 224). 

As shown in \cref{tab:ea}, the top-1 accuracy of ResNet50-1G/2G/3G searched by our method significantly outperforms those of the baseline (SPOS$^\ddag$) by 1.0\%/1.2\%/1.2\% with comparable iter\_time cost, illustrating that the significance of our proposed parallel-subnets training algorithm for efficient and effective supernet training. 
Compared with BCNet~\cite{su2021bcnet}, our pruned ResNet50-2G/3G also exceeds it by 0.4\%/0.6\% with 73\% iter\_time time cost and only 30\% total time cost, further highlighting the supremacy of our presented parallel-subnets training algorithm. 
In addition, our method achieves equivalent performance with the naive serial training technique (i.e., sandwich rule) at a significantly lower expense in time costs (40\% iter\_time and 16\% total time cost).
\begin{table}[!h]
\centering
    \caption{Effect and efficiency analysis of parallel-subnets training algorithm. \#Fw-Bw denotes that the number of executing forward-backward times at each training iteration. Iter\_time means the time cost at each training iteration. $\ddag$ indicates the accuracy is reproduced by us using the same pipeline.}
    \resizebox{0.48\textwidth}{!}{
        \begin{tabular}{l|cccc}
            \toprule
              & \#Fw-Bw & Iter\_time & epochs & Top-1(1G/2G/3G)            \\ \midrule
            SPOS$^\ddag$  \cite{guo2020single}       &  1          & 169ms & 120  & 74.2\% / 76.1\% / 76.7\%               \\              
            BCNet~\cite{su2021bcnet}       &       2      &  303ms & 300 & 75.2\% / 76.9\% / 77.3\%        \\
            AutoSlim$^\ddag$  \cite{yu2019autoslim}     &    4     &  555ms & 300 & 75.1\% / 77.4\% / 77.9\% \\ 
            Ours    &  1   &    221ms & 120 & 75.2\% / 77.3\% / 77.9\%   \\ 
              \bottomrule
    \end{tabular}}
    \label{tab:ea}
\end{table}

\begin{table}[!h]
\centering
    \caption{The effect of our proposed parallel-subnets training algorithm. We adopt MobiNetV2 200M FLOPs for comparison.}
    \resizebox{0.35\textwidth}{!}{
        \begin{tabular}{l|ccc}
            \toprule
             & FLOPs (M) & \#Params (M) & Top-1               \\ \midrule
            EXP1       & 205        & 2.85      &  71.6\%                        \\
            EXP2       & 212        & 3.02       & 71.5\%                   \\
            EXP3      & 210        & 2.93        & 72.4\%   \\ 
            EXP4      & 208        & 3.02        & 72.3\%   \\ 
            PSE-Net      & 213        & 3.03    & 72.7\%      \\ 
              \bottomrule
    \end{tabular}}
    \label{tab:supe}
\end{table}
\subsection{Ablation Study}

{\bf Effect of Our Parallel-subnets Training Algorithm.}
In this part, we verify the efficacy of our proposed parallel-subnets training algorithm by using the MobileNetV2 as unpruned network. 
As illustrated in \cref{impl}, we split the input into $n$ = 4 parts and simulate multiple networks (one largest subnet and three random middle subnets) to execute forward-backward pass in one round at each training iteration. 
Thus, we explore the concrete form of simulating multiple subnets by our parallel-subnets training algorithm. 
Specifically, we implement the following ablation experiments: (1) EXP1: simulating four random middle subnets at each training iteration; (2) EXP2: simulating one smallest subnet and three random middle subnets at each training iteration; (3) EXP3: simulating one largest subnet, one smallest subnet and two random middle subnets at each training iteration.  
(4) EXP4: simulating one largest subnet and two random middle subnets at each training iteration. 
The results are summarized in \cref{tab:supe}. 
We observe that PSE-Net achieves higher performance compared with EXP1, EXP2, EXP3, and EXP4. Compared EXP3, EXP4, and PSE-Net with EXP1 and EXP2, we can find that the largest subnet has a dominant influence on the performance, than other types of subnets.

\begin{table}[!h]
\centering
    \caption{The accuracy performance of the 200M FLOPs searched PSE-Net under various group sizes $K$ of the search space.}
    \resizebox{0.35\textwidth}{!}{
        \begin{tabular}{c|ccc}
            \toprule
            K & FLOPs (M) & \#Params (M) & Top-1               \\ \midrule
            4       & 207        & 2.9      &  71.9\%                       \\
            12       & 214        & 2.8       & 72.1\%                   \\
            20       & 213        & 3.0        & 72.7\%   \\ 
            28      & 212        & 2.9    &72.3\%    \\ 
            36   & 209        & 3.0    & 72.6\%     \\ 
              \bottomrule
    \end{tabular}}
    \label{tab:K}
\end{table}

{\bf Efficiency of Search Space.}
As illustrated in \cref{background}, we partition the channels of each layer into $K$ groups to reduce the search space. 
In this part, we search for the MobileNetV2 with various group sizes $K$ under ImageNet dataset. 
As shown in \cref{tab:K}, our method obtains the best performance at our default value $K=20$. 
Besides, we observe that when $K$ is small, the performance of the searched network improves as $K$ increases. 
This occurs because $K$ influences the size of the search space $\mathcal{A}$, and a larger group size $K$ results in a larger search space, enabling the searched PSE-Net to be closer to the ideal width.
Moreover, performance tends to deteriorate when the group size is too large, indicating that the search space may be too broad for identifying the optimal width. 

\begin{figure}
	\centering
	\subfloat[]{
             \includegraphics[width=0.23\textwidth]{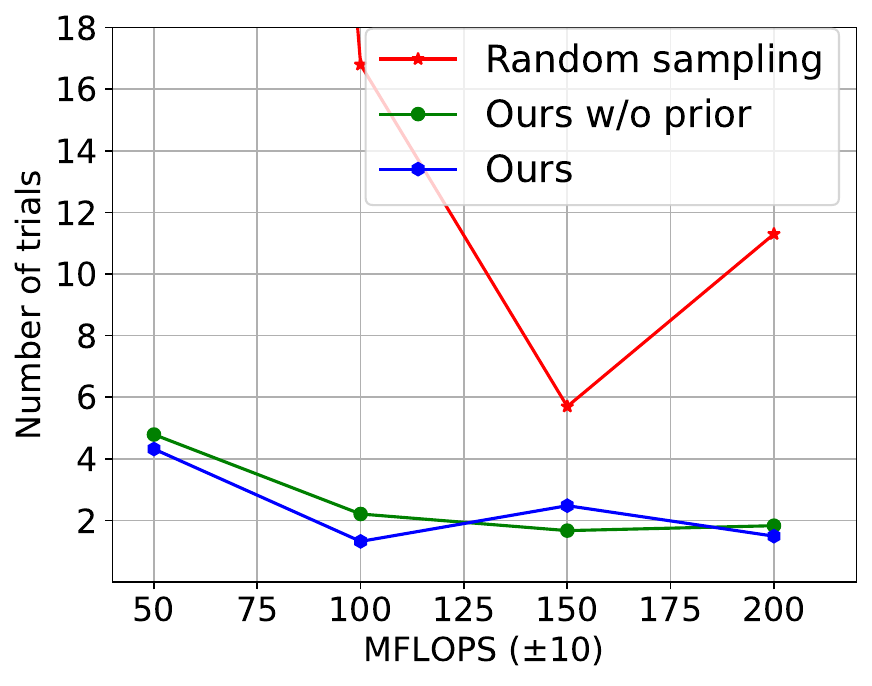}
        }
	\hspace{-1.0em}
	\subfloat[]{
             \includegraphics[width=0.23\textwidth]{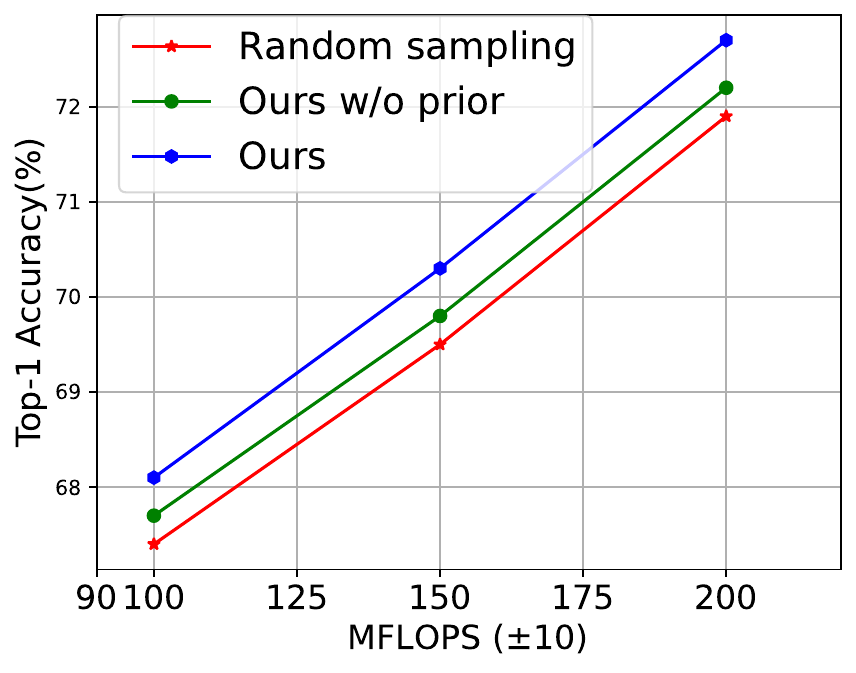}
            }
	\caption{(a) The average number of trials to sample architectures under constraints. (b) Accuracy performance of the searched PSE-Net with different sampling methods.}
	\label{fig:vis_direc}
        \vspace{-1.2em}
\end{figure}

{\bf Effect of Prior-Distributed-Based Sampling Algorithm.}
Our prior-distributed-based sampling algorithm is employed in the search process to sample architectures under different FLOPs constraints. 
In this part, we conduct the following ablation experiments to validate the efficacy of our sampling algorithm: (1) Random sampling: randomly sampling a subnet during the evolutional search process; (2) Ours w/o quality prior information: without using subnet quality (training loss) prior information during sampling, which is also similar to~\cite{wang2021attentivenas}. 
We compare the average number of sampling trials for sampling the targeted architecture under the constraint, as illustrated in \cref{fig:vis_direc} \textcolor{red}{(a)}. 
Due to the long-tail distribution of network architecture, we notice that random sampling poses a challenge to sample subnets that satisfy border resource limitations, whereas our sampling methods with/without quality prior information are both more effective. 
Moreover, as shown in \cref{fig:vis_direc} \textcolor{red}{(b)}, the accuracy performance of subnets searched by our sampling method provides continuous exceptionalism over random sampling and our sampling method without quality prior information, further demonstrating the significance of the prior information of supernet training in assisting the search for the optimal subnet. 

{\bf Effect of Training Supernet with Different Epochs.}
Our method is crucial for searching width as a fundamental performance estimator, which equips the network with the ability to interpolate the accuracy of unsampled subnets. 
In this part, we investigate how much effort should be devoted to supernet training to ensure the search performance. 
Thus, we train our approach with a variety of epochs and leverage the supernets to search MobileNetV2 with 200M FLOPs on the ImageNet dataset. 
The top-1 accuracies of the searched subnets are shown in \cref{tab:epoch}. 
From \cref{tab:epoch}, we can see that the top-1 accuracy of training 90 epochs is 0.9\% better than that of training 30 epochs. 
As the number of epochs continues to increase, there is no significant increase in top1 accuracy. 
Thus, in this work, we train our PSE-Net 90 epochs in total.

\begin{table}[!h]
\centering
    \caption{Accuracy performance of the searched 200M FLOPs PSE-Net under various epochs for supernet training.}
    \resizebox{0.45\textwidth}{!}{
        \begin{tabular}{c|ccc}
            \toprule
            Epochs & FLOPs (M) & \#Params (M) & Top-1               \\ \midrule
            30       & 203        & 2.7      &  71.8\%                       \\
            60       & 205        & 2.8       & 72.4\%                   \\
            90       & 213        & 3.0        & 72.7\%   \\ 
            120      & 208        & 3.0    &72.5\%    \\ 
            150   & 214        & 2.9    & 72.6\%     \\ 
              \bottomrule
    \end{tabular}}
    \label{tab:epoch}
\end{table}

\begin{table}[]
\centering
\caption{Performance of detection task with ResNet50 as the backbone under RetinaNet and Faster RCNN methods.}
 \resizebox{0.475\textwidth}{!}{\begin{tabular}{c|ccc}
\toprule
Methods   & Original (4G)  & BCNet (2G)    & PSE-Net (2G)  \\
RetinaNet~\cite{lin2017focal}   & 36.4\%         & 35.4\%                   & 35.9\%             \\
Faster RCNN~\cite{ren2015faster} & 37.3\%         & 36.3\%                   & 36.8\%             \\ 
\bottomrule
\end{tabular}}
\label{detection}
\end{table}

\begin{table}[]
\centering
\caption{Performance of semantic segmentation task with ResNet50 as the backbone under Deeplabv3 method.}
 \resizebox{0.475\textwidth}{!}{\begin{tabular}{c|ccc}
\toprule
Method   & Original (4G)  &  BCNet(2G) & PSE-Net (2G)  \\
Deeplabv3~\cite{chen2017rethinking} & 78.58\%  &  77.82   & 78.12\%          \\
\bottomrule
\end{tabular}}
\label{segman}
\end{table}

\subsection{Transferability to Downstream Tasks}
Then, we applied the pruned PSE-Net to tasks of object detection and semantic segmentation based on frameworks MMDetection~\cite{mmdetection} and MMSegmentation~\cite{mmseg2020}, respectively.

{\bf Object Detection.}
In this part, we evaluate PSE-Net on object detection task.
Experiments are performed using the COCO dataset, which includes 118K training images and 5K validation images of a whole of 80 classes. 
We take the searched 2G FLOPs ResNet50 subnets as the backbone under the two typical detection frameworks: RetinaNet~\cite{lin2017focal} and Faster R-CNN~\cite{ren2015faster}. 
Following the popular MMDetection~\cite{mmdetection} toolbox, we train both RetinaNet and Faster R-CNN for 12 epochs, with a mini-batch of 16/16 and an initial learning rate of 0.01/0.02 for RetinaNet/Faster R-CNN.
The learning rate is divided by 10 after 8 and 11 epochs. 
We employ SGD optimizer with a weight decay of 1e-4. 
During the training and testing stages, the shorter side of input photos are shrunk to 800 pixels along the shorter dimension. 
The longer side will maintain the image ratio within 1333 pixels. 
In the training stage, we only leverage random horizontal flipping for data augmentation. 
We report the results in terms of AP in \cref{detection}.  
PSE-Net (2G) surpasses the current state-of-the-art pruning algorithm BCNet~\cite{su2021bcnet} by 0.5\% and 0.5\%, respectively, in both the RetinaNet and Faster R-CNN frameworks, demonstrating its supremacy. 

{\bf Semantic Segmentation.}
We also evaluate the searched 2G FLOPs ResNet50 on the semantic segmentation task with the Cityscapes dataset. 
The Cityscapes dataset is a large-scale urban street scene segmentation dataset. 
It contains high-resolution street scene images from 50 cities across the world, with a pixel-level annotation of 19 semantic classes, including vehicles, buildings, road surfaces, vegetation, pedestrians, cyclists, and more. 
We select Deeplabv3~\cite{chen2017rethinking} as a basic method to validate our pruned 2G FLOPs ResNet50. 
The networks are trained for 40K iterations by applying SGD, with an initial learning rate of 0.01 and weight decay of 4e-5. 
We adopt the poly learning rate schedule with $\gamma$=0.9 and each mini-batch of 16. 
We resize and randomly crop images to 769 $\times$ 769 for training. 
During the testing stage, images are resized to the shorter side of 1025 pixels. 
We report mIoU of PSE-Net in single scale testing. 
The results are summarized in \cref{segman}. 
We observe that PSE-Net(2G) achieves 78.12\% mIoU, which surpasses the current state-of-the-art pruning algorithm BCNet(2G) than 0.3\% and is slightly lower than the original model with a decrease of 2G FLOPs, further demonstrating the superiority of PSE-Net. 

\subsection{Latency of PSE-Net}
Since there is a gap between FLOPs and the hardware latency, we compare the latency of PSE-Net with that of the baseline (the original ResNet50) and other state-of-the-art methods, as shown in \cref{latency}. 
The latency of these methods are measured on a Tesla A100 GPU with input shape of data as (64, 3, 224, 224). 
PSE-Net surpasses the original ResNet50 model by 5.75/8.35/11.95 units under 3G/2G/1G FLOPs settings, respectively. Compared with DMCP, our 3G/2G/1G FLOPs ResNet50 models all achieve better performance. Compared with BCNet, our 3G/1G FLOPs ResNet50 models perform more efficiently on hardware, while the 2G FLOPs setting also achieves comparable result, further demonstrating the efficacy of PSE-Net. 
\begin{table}[]\centering
    \caption{Latency of PSE-Net compared with the baseline and other state-of-the-art methods. To make the results reliable and stable, we warm up the networks and measure their latency multiple times and calculate their means.\label{latency}}
    \resizebox{0.45\textwidth}{!}{
    \begin{tabular}{l|cccc}
    \toprule
  & Orignal(4G) & 3G & 2G &  1G \\ \midrule
    ResNet50~\cite{he2016deep}     & 25.50ms         & -    & -       & - \\
    DMCP~\cite{guo2020dmcp} & -           & 23.05ms      & 20.05ms       & 14.45ms \\
    BCNet~\cite{su2021bcnet}  & -         & 22.35ms    & 17.05ms       &  15.05ms\\
    PSE-Net     & -         & 19.75ms    & 17.15ms       & 13.55ms \\
    \bottomrule
    \end{tabular}}

\end{table}

\section{Conclusion}
In this work, we introduce PSE-Net, a novel parallel-subnets estimator dedicated to efficient yet effective channel Pruning. 
PSE-Net is equipped with the proposed training algorithm, the parallel-subnets training algorithm. 
Under this algorithm, PSE-Net can simulate forward-backward of multiple sampled subnets in one round, improving the training efficiency of the supernet while arming the network with the capability of interpolating the accuracy of unsampled subnets. 
Over the trained PSE-Net, we devise a prior-distributed-based sampling algorithm to boost the performance of the classical evolutionary search. 
Such an algorithm employs the prior knowledge of supernet training to aid in the search for optimal subnets while addressing the difficulty of finding samples that satisfy resource limitations due to the long-tail distribution of network setup. 
Extensive experiments demonstrate PSE-Net outperforms previous state-of-the-art channel pruning methods on the ImageNet dataset and downstream tasks. 

\bibliographystyle{cas-model2-names}
\bibliography{egbib}
\vfill
\end{document}